\documentclass[conference]{IEEEtran}
\IEEEoverridecommandlockouts
\usepackage{cite}
\usepackage{hyperref}
\usepackage{amsmath,amssymb,amsfonts}
\usepackage{graphicx}
\usepackage{textcomp}
\usepackage{multirow}
\usepackage{xcolor}
\usepackage{listings}
\usepackage{tcolorbox}
\usepackage{algpseudocode}
\tcbuselibrary{breakable}
\def\BibTeX{{\rm B\kern-.05em{\sc i\kern-.025em b}\kern-.08em
    T\kern-.1667em\lower.7ex\hbox{E}\kern-.125emX}}

\lstdefinestyle{mystyle}{
  basicstyle=\ttfamily\small,
  columns=fullflexible,
  breaklines=true,
  frame=none,
  xleftmargin=2em,
  xrightmargin=2em
}

\begin{document}

\title{Efficient Intent-Based Filtering for Multi-Party Conversations Using Knowledge Distillation from LLMs}

\author{
\IEEEauthorblockN{Reem Gody}
\IEEEauthorblockA{\textit{Microsoft AI} \\
regody@microsoft.com}
\and
\IEEEauthorblockN{Mohamed Abdelghaffar}
\IEEEauthorblockA{\textit{Microsoft AI} \\
moabdelg@microsoft.com}
\and
\IEEEauthorblockN{Muhammad Jabreel}
\IEEEauthorblockA{\textit{Microsoft AI} \\
mjabreel@microsoft.com}
\and
\IEEEauthorblockN{Ahmed Y. Tawfik}
\IEEEauthorblockA{\textit{Microsoft AI} \\
atawfik@microsoft.com}
}

\maketitle

\begin{abstract}
Large language models (LLMs) have showcased remarkable capabilities in conversational AI, enabling open-domain responses in chat-bots, as well as advanced processing of conversations like summarization, intent classification, and insights generation. However, these models are resource-intensive, demanding substantial memory and computational power. To address this, we propose a cost-effective solution that filters conversational snippets of interest for LLM processing, tailored to the target downstream application, rather than processing every snippet. In this work, we introduce an innovative approach that leverages knowledge distillation from LLMs to develop an intent-based filter for multi-party conversations, optimized for compute power constrained environments. Our method combines different strategies to create a diverse multi-party conversational dataset, that is annotated with the target intents and is then used to fine-tune the MobileBERT model for multi-label intent classification. This model achieves a balance between efficiency and performance, effectively filtering conversation snippets based on their intents. By passing only the relevant snippets to the LLM for further processing, our approach significantly reduces overall operational costs depending on the intents and the data distribution as demonstrated in our experiments.
\end{abstract}


\section{Introduction}

The emergence of LLMs has shown great potential in effectively handling complex conversational tasks. These models leverage vast amounts of data and sophisticated architectures to understand and predict user intentions in conversations with impressive accuracy \cite{liu-etal-2024-lara,mao-etal-2023-large}. LLMs can understand implicit intents in long conversations and generate accurate responses. They can also categorize conversations based on their overall intent. For example, \cite{Shah2023UsingLL} utilized LLMs with human-in-the-loop to generate, validate, and apply user taxonomies on Bing chat conversations, which represent good examples for machine-user interactions. LLMs can manage open-domain multi-party conversations in a way that was not possible with older technologies, which were more focused on task-oriented dialogues due to their limitations.

These capabilities have opened up new possibilities for applications that can handle large volumes of conversations more effectively. One potential application is the real-time processing of meeting transcripts to generate insights or identify actionable items once the meeting concludes. In this scenario, the LLM would analyze multi-party conversations, using the context to highlight key points and assign tasks to different participants. Another application involves analyzing interactions between machines and users to trigger tools or proactive actions when developing virtual assistants, conversational search systems, or chat-bots. In each of these cases, it's evident that many substantial parts of the conversation, like chit-chat, or side conversations, may be irrelevant to the primary task the LLM needs to perform, making it cost inefficient to process these irrelevant exchanges.

\begin{figure}[h!]
\centerline{\includegraphics[width=\linewidth]{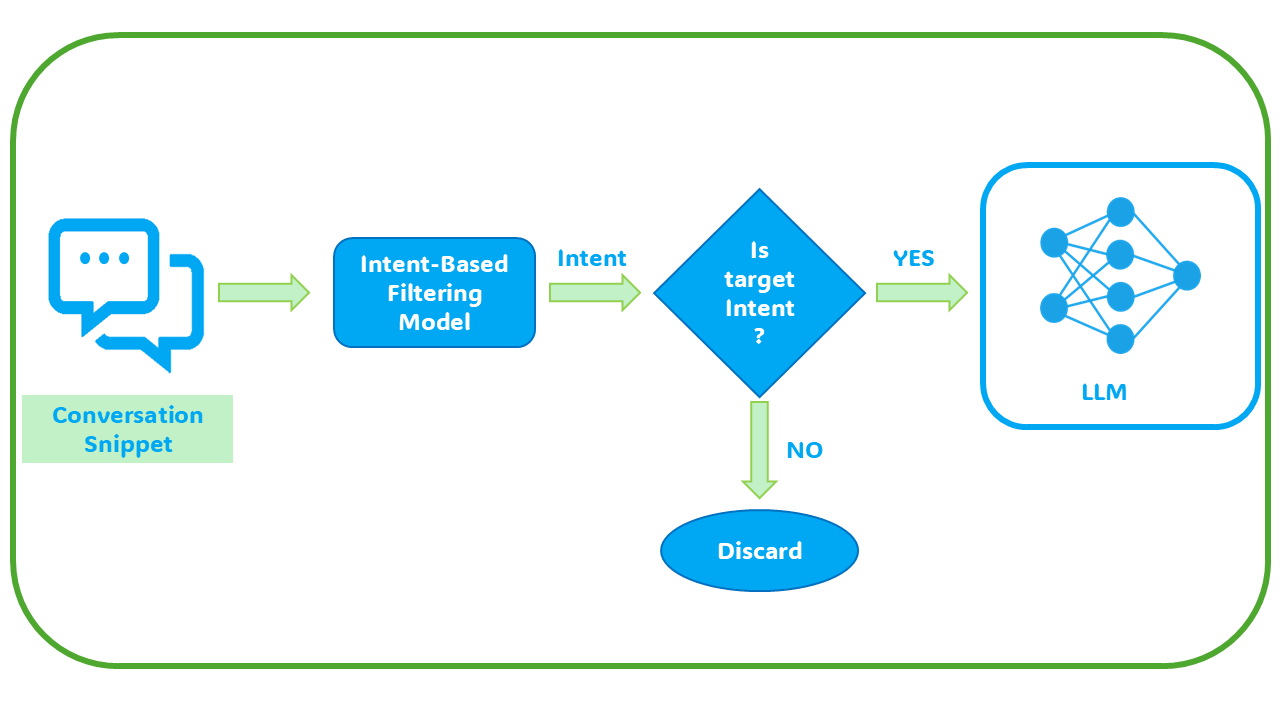}}
\caption{An intent-based filtering model is used to filter irrelevant conversation snippets.}
\label{fig:idea}
\end{figure}

This brings up the question of whether we can filter conversational snippets of interest to pass to the LLM, rather than sending the full conversation to the LLM for further analysis and insights generation. In particular, we consider using the LLM's conversational understanding capabilities and explore the potential of distilling the LLM's knowledge to create an intent-based filtering model. This model checks whether a conversation snippet contains the intents relevant to a particular downstream application before passing it to the LLM for further processing as illustrated in Figure \ref{fig:idea}. This approach reduces the token count processed by the LLMs, thereby lowering the overall system cost. These lightweight intent filtering models must balance efficiency and performance, ensuring high quality without requiring powerful hardware. To our knowledge, this is the first work focusing on intent-based conversation filtering in multi-party conversations with an emphasis on resource efficiency. We propose a novel approach that filters conversation snippets based on their intents while maintaining a low computational footprint. Our main contributions are: 
\begin{itemize} 
\item \textbf{Custom dataset creation}: We utilize black box knowledge distillation to build a specialized dataset for multi-party conversational scenarios using the LLM as a teacher to generate and label conversations. These conversations are designed to capture the diversity and complexity of real-life interactions to train and evaluate our intent-based filtering model. 
\item \textbf{Lightweight model architecture}: We demonstrate our approach on the MobileBERT model \cite{mobilebert} since it is light-weight and optimizes computational efficiency, making it suitable for deployment on edge devices without compromising performance. 
\item \textbf{Real-world applicability}: We demonstrate the practical application of our approach in various settings, such as team meetings and collaborative projects involving multiple participants, where our light-weight model can help filter conversations based on their intents, and only pass the conversation snippets of interest for further processing using an LLM, thereby reducing the token count and the overall operational cost. 
\end{itemize}

\section{Related Work}

In this section, we discuss the related work in building conversational datasets, and distilling the knowledge from LLMs through data annotation.

\subsection{Conversational Data}
To support a diverse array of conversational AI LLM-based applications, our intent-based filtering model must function effectively in open-domain, multi-party conversations. This necessitates the collection of substantial amounts of conversational data labeled with the relevant intents for fine-tuning the model. Historically, most pre-LLM research in Conversational AI has focused on either conversational search or task-oriented dialogue systems, resulting in the predominance of domain-specific multi-turn dialogue data \cite{hemphill-etal-1990-atis}, \cite{busso2008iemocap}, \cite{henderson2014second}, \cite{henderson2014third}, \cite{williams2014dialog}, \cite{williams2016dialog}, \cite{el-asri-etal-2017-frames}, \cite{eric-etal-2017-key}, \cite{Shah2018BuildingAC}. Although initiatives such as MultiWOZ \cite{zang2020multiwoz} have extended to multi-domain settings, they remain limited in scope and typically involve two-speaker interactions. A few datasets, such as DailyDialog \cite{li-etal-2017-dailydialog} and those based on social media like Twitter \cite{ritter-etal-2010-unsupervised}, have attempted to cover open-domain settings but lack the necessary annotations for conversation-level intent classification. Similarly, datasets derived from movies \cite{lison2016opensubtitles2016}, \cite{Danescu-Niculescu-Mizil+Lee:11a} and TV series \cite{yang2019friendsqa} present plausible candidates for multi-party multi-turn natural conversational data, yet they are deficient in relevant annotations and are not entirely realistic due to the scripted nature of the content. The AMI meeting corpus \cite{AMIcorpus} also represents a valuable resource for multi-party interactions, containing a significant number of scenario-driven meetings, but it is only annotated with dialogue acts and lacks conversation-level intent annotations. This presents two primary challenges: the need to acquire more multi-party general domain conversations for fine-tuning the intent-based filter model, and the necessity to provide conversation-level annotations for the target intents of interest across all available conversational corpora.
\subsection{LLMs For Synthetic Conversational Data Generation}
Several works have discussed open-domain conversational data generation as \cite{kim-etal-2023-soda}, \cite{chen-etal-2023-places}, and \cite{myat2024framework}. Similar to these works, we leverage LLMs to generate more representative data for open-domain multi-party conversations.
\subsection{Knowledge Distillation using LLM-Based Annotations}
Due to the absence of conversation-level intent labels in existing conversational data, it is necessary to employ human annotators to label conversations for specific application needs. However, this process is both expensive and time-consuming. Several studies have explored the use of LLMs for annotating textual data in various NLP tasks as \cite{ding2023gpt3gooddataannotator}, and \cite{pangakis2023automatedannotationgenerativeai}. He et al. \cite{he2024annollmmakinglargelanguage} argue that techniques like chain of thought and prompting LLMs to provide explanations for annotations can match or surpass crowd-sourced annotations. 

Pangakis et al. \cite{pangakis2024knowledgedistillationautomatedannotation} demonstrate that LLM knowledge can be distilled into smaller BERT-based models \cite{Devlin2019BERTPO} for several classification tasks. They utilize LLM-generated annotations for textual classification datasets in the supervised fine-tuning of BERT-based models for 14 classification tasks, showing that models fine-tuned on LLM generated labels perform comparably to those fine-tuned on human labels. Similarly, we use LLMs to generate intent labels for conversations and employ supervised fine-tuning for the MobileBERT model \cite{mobilebert} to build an intent-based conversation filter. Given the unique nature of conversational data, our labeling pipeline leverages specific characteristics of conversations such as the presence of turns.

\section{Methodology}
 Our approach relies on distilling LLM capabilities to a smaller classifier that is faster, resource efficient and easy to run on a large scale. This enables fast and cost-effective filtering of conversational data compared to the much slower and more expensive LLM-calls. To achieve this, we go through the following stages of our solution:
 \begin{enumerate}
\item \textbf{Data Collection}: This involves collecting publicly available conversational data, as well as utilizing the LLM for generating synthetic conversations.
\item \textbf{Data Labeling}: Since the collected conversational data lacks the intent labels, we utilize the LLM for generating custom annotations depending on the target scenarios.
\item \textbf{Data Augmentation}: This involves sampling potentially overlapping conversation snippets from the original conversations to enhance the robustness of the model to various conversation lengths.
\item \textbf{Model Fine-Tuning}: We fine-tune a resource-efficient model for the multi-label classification task using the conversational snippets annotated with the target intents.
\end{enumerate}

We demonstrate the advantages of our solution on two example scenarios that are likely to be useful for multiple downstream applications: “action-triggering" and “information-seeking" conversations. We define action-triggering conversations as conversations that prompt action or involve promises, tasks, reminders, or commitments made by the user. We define information-seeking conversations as those involving inquiries about a topic, or demonstrating lack of clear knowledge, or curiosity.
\subsection{Data Collection and Synthesis}
To serve both target scenarios, we collect conversations that contain implicit or explicit calls to action, as well as those which express information need. In addition, we collect general chit-chats, daily conversations and irrelevant conversations to our scenarios of interest to account for negative examples in the training data.
\begin{itemize}
\item \textbf{Natural Conversational Data Collection}
To create a rich and diverse dataset covering a wide range of domains, we collect data from multiple publicly available conversational datasets. We select the FriendsQA dataset \cite{yang2019friendsqa} and the Movie Dialogue dataset \cite{Danescu-Niculescu-Mizil+Lee:11a} for open-domain multi-party conversations, and the AMI Meeting Corpus \cite{AMIcorpus} for multi-party conversations in meeting settings. To enhance the representation of information-seeking conversations, we use the Question Answering in Context (QuAC) dataset \cite{choi-etal-2018-quac} and the Conversational Curiosity Corpus \cite{rodriguez2020curiosity}, as well as a sampled subset of the Natural Questions dataset \cite{kwiatkowski2019natural}. The MultiWOZ dataset \cite{zang2020multiwoz}, a task-oriented dialogue dataset covering multiple domains, is chosen as it has a good representation of both action-triggering and information-seeking conversations. We also increase the representation of the action-triggering conversations by adding other task oriented dialogue datasets like the “Taxi conversations" \cite{Li2018MicrosoftDC}. Additionally, we include datasets like the Tennis Interviews dataset \cite{fu2016tie}, featuring post-game press conferences with tennis players, the Campsite Negotiations dataset (CaSiNo) \cite{chawla2021casino}, and the Intelligence Squared Debates dataset (IQ2) \cite{zhang2016conversational}, which features argumentative and persuasive dialogues. 

Alongside these public datasets, we utilize an in-house smart assistant dataset that includes single-turn interactions between a user and an assistant agent for tasks like setting reminders, alarms, calendar events, etc., to increase the representation of the action-triggering intent. We utilize this data as is, and sample part of it to be used as a seed for synthetic data generation.

\item \textbf{Synthetic Conversational Data Generation}
 To increase the representation of the target intents, in the conversational data used to fine-tune the intent classification model, we resort to generating synthetic conversational data that contain implicit and explicit calls to action as well as inquiries or need for information. In addition, we generate general conversational data spanning multiple topics to represent natural conversations that are likely to occur between multiple people on daily basis in order to enhance the presence of general multi-party conversations in the training data.
We experiment with two approaches:
\begin{enumerate}
\item \textbf{\textit{Data Generation Based on User Queries}}

We leverage the GPT-4o model to synthesize conversations that are conditioned on input queries that are extracted from natural datasets. This allows the synthetic conversations to be diverse and closer to the scenarios of interest. To generate action-triggering conversations, we use the assistant commands from an in-house smart-assistant dataset as input queries to the generation prompt. We refer to the generated dataset using this approach as “Synthetic Multi-turn Assistant" dataset. To generate information-seeking conversations, we leverage Google’s natural questions dataset \cite{kwiatkowski2019natural} and generate conversations revolving around the topics of the questions. We refer to this generated dataset as “Synthetic Information Seeking" dataset.
\item \textbf{\textit{Persona-Based Synthetic Data Generation}}

Similar to various works that leveraged LLMs for conversational data generation \cite{chen-etal-2023-places, jandaghi-etal-2024-faithful}, we leverage GPT-4o for conversation synthesis. To make these conversations diverse, we build a multi-agent system using the AutoGen framework \cite{Wu_AutoGen_Enabling_Next-Gen_2024}. We first use GPT-4o to create personas for the different agents and a conversation starter. The persona of each agent has the name, qualities and speech style for that agent. A group of agents is then spawned, where each agent is initialized with their profile and then the group chat manager initiates a conversation between the agents using the generated conversation starter. We refer to the dataset generated using this approach as the “Synthetic Situations" dataset.
\end{enumerate}
\end{itemize}
\subsection{Data Labeling}

Several studies have harnessed the capabilities of LLMs to generate high-quality dataset annotations, as demonstrated by \cite{ding2023gpt3gooddataannotator} and \cite{pangakis2023automatedannotationgenerativeai}. AnnoLLM \cite{he2024annollmmakinglargelanguage} illustrated that with adequate guidance, example demonstrations, and prompting the LLM to provide explanations for its annotations, the performance can surpass or match that of crowd-sourced annotations.

In our approach, we establish a distinct labeling pipeline for each intent. This involves providing a comprehensive definition of the intent, supplemented with both positive and negative examples, along with explanations for why an example is classified as positive or negative for that intent. We utilize the GPT-4o model in our experiments, prompting it to explain its labeling decisions to enhance the quality of the annotations.

We employ binary indicator labels to denote whether a conversation contains a specific intent. Our objective is to generate multiple intent labels for each conversation. To achieve this, we label each turn of the conversation with all target intents. The conversation-level labels are then derived by aggregating the turn-level labels. A conversation is assigned a positive label for a particular intent if at least one turn within the conversation is labeled positively for that intent.

This turn-level labeling enables the extraction of various segments from each conversation, with labels for these segments being derived through the aggregation of turn-level labels. This process is crucial for our data augmentation strategy.

\subsection{Data Augmentation}

We implement an online data augmentation strategy that dynamically generates additional training samples from existing conversations using a rolling window technique. This technique involves selecting random window sizes within predefined minimum and maximum turn limits. From each conversation, we extract up to $x$ additional segments of these random sizes, starting from random positions within the conversation, such that $x$ is a configurable hyperparameter. This method is applied to a percentage of the training batches using a configurable probability to introduce diverse conversation segments without over-fitting. This approach enables the model to adapt to conversations with varying turn lengths, ensuring that model decisions are based solely on textual content, thereby mitigating biases related to conversation length and enhancing the model's robustness and performance.

\begin{table*}[t!]
\centering
\caption{\centering Test Datasets' Performance Metrics}
\begin{tabular}{|l|r|r|r|r|r|r|}
\hline
\textbf{\multirow{2}{*}{Dataset}} & \multicolumn{3}{c|}{\textbf{Action-Triggering}} & \multicolumn{3}{c|}{\textbf{Information-Seeking}} \\
\cline{2-7}
 & \textbf{Precision} & \textbf{Recall} & \textbf{F1-score} & \textbf{Precision} & \textbf{Recall} & \textbf{F1-score} \\
\hline
AMI - 1 min chunks & 0.72 & 0.78 & 0.75 & 0.71 & 0.73 & 0.72 \\
\hline
Synthetic Situations & 0.85 & 0.99 & 0.92 & 0.56 & 0.56 & 0.56  \\
\hline
Movie Corpus & 0.86 & 0.70 & 0.77 & 0.60 & 0.53 & 0.56  \\
\hline
Synthetic Multi-turn Assistant & 0.99 & 0.99 & 0.99 & 0.46 & 0.52 & 0.49  \\
\hline
In-house Assistant Commands & 0.99 & 0.90 & 0.95 & 0.43 & 0.57 & 0.49  \\
\hline
MultiWOZ & 0.98 & 0.98 & 0.98 & 0.97 & 0.99 & 0.98  \\
\hline
Curiosity Dialogues - 0 & 0.62 & 0.48 & 0.54 & 0.99 & 0.99 & 0.99  \\
\hline
Curiosity Dialogues - 1 & 0.63 & 0.44 & 0.52 & 0.99 & 0.99 & 0.99  \\
\hline
Balanced Test & 0.90 & 0.91 & 0.90 & 0.95 & 0.95 & 0.95  \\
\hline
\end{tabular}

\label{tab:test_dataset_performance}
\end{table*}

\begin{table*}[t!]
\centering
\caption{Reduction in Tokens Consumed by the LLM}

\begin{tabular}{|l|r|p{1.6cm}|p{1.3cm}|p{1.6cm}|p{1.3cm}|p{1.6cm}|p{1.3cm}|}

\hline
 \multicolumn{2}{|c}{\textbf{Intent}} & \multicolumn{2}{|c|}{\textbf{Action-Triggering}} & \multicolumn{2}{|c|}{\textbf{Information-Seeking}} & \multicolumn{2}{|c|}{\textbf{Any}} \\ \hline
\textbf{Test Set} & \textbf{Total tokens} & \textbf{Expected Reduction \%} & \textbf{Actual Reduction \%} &  \textbf{Expected Reduction \%} & \textbf{Actual Reduction \%} & \textbf{Expected Reduction \%} & \textbf{Actual Reduction \%} \\  \hline
Balanced Test & 485642 & \multicolumn{1}{r|}{48.29} & \multicolumn{1}{r|}{47.21} & \multicolumn{1}{r|}{28.75} & \multicolumn{1}{r|}{27.97} &  \multicolumn{1}{r|}{13.19} & \multicolumn{1}{r|}{9.07} \\ \hline
AMI - 1 min chunks & 154665 & \multicolumn{1}{r|}{50.05} & \multicolumn{1}{r|}{46.08} &  \multicolumn{1}{r|}{59.01} & \multicolumn{1}{r|}{57.52} &  \multicolumn{1}{r|}{30.42} & \multicolumn{1}{r|}{25.27} \\ \hline
Movie Corpus & 12806 &  \multicolumn{1}{r|}{34.54} & \multicolumn{1}{r|}{47.92}  & \multicolumn{1}{r|}{87.25} & \multicolumn{1}{r|}{90.28} &  \multicolumn{1}{r|}{32.55} & \multicolumn{1}{r|}{45.53} \\ \hline
Synthetic Situations  & 38801 &  \multicolumn{1}{r|}{31.55} & \multicolumn{1}{r|}{22.43}  & \multicolumn{1}{r|}{87.00} & \multicolumn{1}{r|}{87.24} &  \multicolumn{1}{r|}{31.01} & \multicolumn{1}{r|}{21.90} \\ \hline
\end{tabular}
\label{tab:token_count_table}
\end{table*}
\subsection{Model Fine-tuning}
We use the curated datasets to fine-tune a small pre-trained model for the multi-label classification task, while applying the previously described online data augmentation approach. We split the conversations that don't fit within the context window of the model into multiple segments to fit, and fine-tune all the parameters of the model. To formulate multiple speaker turns of a conversation into a single input we join them using the ‘separate token’ (aka SEP token), and drop the speaker information. Since each of the speaker turns have their own label, these labels are aggregated into a single label to be presented as one input sample during fine-tuning. We aggregate the labels via a logical OR since having any positive occurrence of the intent for any of the conversation turns automatically means that the entire input sample should be labeled with that intent.

\section{Experimental Setup}
To build the intent-based filtering model, we fine-tune the MobileBERT model \cite{mobilebert} for the multi-label intent classification task using Binary Cross Entropy loss. We use two binary labels mapping to the two scenarios: action-triggering and information-seeking conversations. Tables \ref{tab:train_dataset_summary} and \ref{tab:val_dataset_summary} in Appendix \ref{appendix:DataCollection} have a detailed breakdown of the training and validation datasets used in our experiments. In our best experiment, We fine-tune the model on a single NVIDIA V100 GPU with 16 Gigabytes of VRAM for five epochs using a learning rate of $2 \cdot 10^{-5}$ and a batch size of 24 samples per batch. To reduce over-fitting, We apply L2 regularization and set the regularization term to $1 \cdot 10^{-2}$. For online data augmentation, we use a rolling window that ranges from one to five turns to simulate the different chunking of the conversations. We apply the rolling window technique to 50\% of the training batches, and sample up to 2 conversation segments from each conversation. We compute the recall, precision, and F1 score for each intent label on the validation data every 500 steps, and select the best checkpoint overall.

\section{Evaluation}
We conduct the evaluation of the fine-tuned MobileBERT model on several held-out sets that underwent the same labeling pipeline as the training data. Our test suite includes a sampled subset of the Movies Dialogue dataset \cite{Danescu-Niculescu-Mizil+Lee:11a} and the Synthetic Situations dataset, which represent general multi-party conversations. In addition,  We utilize the official test subsets of the Curiosity Dialogues \cite{rodriguez2020curiosity}, AMI \cite{AMIcorpus}, and MultiWOZ \cite{zang2020multiwoz} datasets, but label them using our data labeling pipelines for the target intents. The Curiosity Dialogues dataset effectively represents information-seeking conversations and has two official test subsets, one of them has the same distribution as the training data (1), and the other one is totally unseen and is utilized for zero-shot evaluations (0). The MultiWOZ dataset captures both action-triggering and information-seeking conversations. The AMI dataset is a suitable candidate for meeting scenarios. Additionally, we evaluate held-out sets from our in-house smart assistant data, and the Synthetic Multi-turn Assistant dataset. The balanced test set is sampled from AMI, Synthetic Situations, Movies, Synthetic Multi-turn Assistant, MultiWOZ, and Curiosity Dialogues to ensure a reasonable distribution of the intent labels. Table \ref{tab:test_dataset_summary} in the Appendix \ref{appendix:DataCollection} provides more details on the test sets used.

In Table \ref{tab:test_dataset_performance}, we report the precision, recall, and F1 score for the positive occurrence of each label for the best checkpoint. As shown in the table, we achieve an F1 score of 0.95 for information-seeking conversations and 0.9 for action-triggering conversations on the balanced test set.

For the reported F1 scores, we analyze the reduction in token count achieved by only sending conversations with action-triggering or information-seeking intents to the LLMs. This analysis is conducted on different datasets with varying intent distributions, as the reduction is highly dependent on the intent label distribution in the test sets. We also compare the expected reduction, based on reference labels, with the actual reduction, based on the intent model's predictions. Equation \ref{eq:actual_reduction} shows how to compute the actual reduction in token count, where “Prediction Tokens" is the number of tokens in the conversations for which the model predicts the intent label. Similarly, equation \ref{eq:expected_reduction} shows how to compute the expected reduction in token count, such that “Reference Tokens" is the number of tokens in the conversations that have the intent label. As demonstrated in Table \ref{tab:token_count_table}, the model significantly reduces the tokens consumed by the LLM. For example, in the AMI test, which is a good example of meeting transcriptions, we achieve a reduction of 46.08\% at an F1 score of 0.75 when sending only action-triggering conversations to the LLM, a reduction of 57.52\% at an F1 score of 0.72 when sending only information-seeking conversations, and a reduction of 25.27\% when sending any conversation that triggers either intent to the LLM for further processing.

\begin{equation}
\label{eq:actual_reduction}
\text{Actual Reduction} = 100 - \left( \frac{\text{Prediction Tokens}}{\text{Total Tokens}} \right)
\end{equation}

\begin{equation}
\label{eq:expected_reduction}
\text{Expected Reduction} = 100 - \left( \frac{\text{Reference Tokens}}{\text{Total Tokens}} \right)
\end{equation}

\section{Conclusion}
 In conclusion, our proposed approach addresses the significant resource demands of LLMs in conversational AI by introducing an efficient, intent-based filtering mechanism. By leveraging knowledge distillation from LLMs, we develop a model that effectively classifies multi-party conversation snippets based on their intents, optimizing for resource-constrained environments. 
The evaluation results demonstrate the model's capability to accurately classify information-seeking and action-triggering conversations, achieving high F1-scores on various test sets. While we demonstrate our approach on two specific target intents, it can be easily extended to other intents by collecting and generating relevant conversational datasets and annotating them with the desired intents using the approaches we described. Furthermore, our proposed approach significantly reduces the token count sent to LLMs, thereby lowering the operational costs. Overall, our intent-based filtering model not only enhances the efficiency of LLM-based applications in processing conversational data but also ensures that only relevant snippets are forwarded to the LLM for further analysis. This innovation paves the way for more cost-effective and scalable solutions in conversational AI, making advanced language models more accessible and practical for a wide range of applications.

\section*{Acknowledgment}
We used Microsoft Copilot for revising, and enhancing the writing of all the sections in the paper.

\bibliographystyle{unsrt} 
\bibliography{references} 

\begin{thebibliography}{10}

\bibitem{liu-etal-2024-lara}
Junhua Liu, Tan~Yong Keat, Bin Fu, and Kwan~Hui Lim.
\newblock {LARA}: Linguistic-adaptive retrieval-augmentation for multi-turn
  intent classification.
\newblock In {\em Proceedings of the 2024 Conference on Empirical Methods in
  Natural Language Processing: Industry Track}, pages 1096--1106. Association
  for Computational Linguistics, November 2024.

\bibitem{mao-etal-2023-large}
Kelong Mao, Zhicheng Dou, Fengran Mo, Jiewen Hou, Haonan Chen, and Hongjin
  Qian.
\newblock Large language models know your contextual search intent: A prompting
  framework for conversational search.
\newblock In {\em Findings of the Association for Computational Linguistics:
  EMNLP 2023}, pages 1211--1225. Association for Computational Linguistics,
  December 2023.

\bibitem{Shah2023UsingLL}
C.~Shah, Ryen~W. White, Reid Andersen, Georg Buscher, Scott Counts, Sarkar
  Snigdha~Sarathi Das, Ali Montazer, Sathish Manivannan, Jennifer Neville,
  Xiaochuan Ni, N.Kasturi Rangan, Tara Safavi, Siddharth Suri, Mengting Wan,
  Leijie Wang, and Longfei Yang.
\newblock Using large language models to generate, validate, and apply user
  intent taxonomies.
\newblock {\em ArXiv}, abs/2309.13063, 2023.

\bibitem{mobilebert}
Zhiqing Sun, Hongkun Yu, Xiaodan Song, Renjie Liu, Yiming Yang, and Denny Zhou.
\newblock {M}obile{BERT}: a compact task-agnostic {BERT} for resource-limited
  devices.
\newblock In {\em Proceedings of the 58th Annual Meeting of the Association for
  Computational Linguistics}, pages 2158--2170. Association for Computational
  Linguistics, July 2020.

\bibitem{hemphill-etal-1990-atis}
Charles~T. Hemphill, John~J. Godfrey, and George~R. Doddington.
\newblock The {ATIS} spoken language systems pilot corpus.
\newblock In {\em Speech and Natural Language: Proceedings of a Workshop Held
  at Hidden Valley}, 1990.

\bibitem{busso2008iemocap}
Carlos Busso, Murtaza Bulut, Chi-Chun Lee, Abe Kazemzadeh, Emily Mower, Samuel
  Kim, Jeannette~N Chang, Sungbok Lee, and Shrikanth~S Narayanan.
\newblock Iemocap: Interactive emotional dyadic motion capture database.
\newblock {\em Language resources and evaluation}, 42:335--359, 2008.

\bibitem{henderson2014second}
Matthew Henderson, Blaise Thomson, and Jason~D Williams.
\newblock The second dialog state tracking challenge.
\newblock In {\em Proceedings of the 15th annual meeting of the special
  interest group on discourse and dialogue (SIGDIAL)}, pages 263--272, 2014.

\bibitem{henderson2014third}
Matthew Henderson, Blaise Thomson, and Jason~D Williams.
\newblock The third dialog state tracking challenge.
\newblock In {\em 2014 IEEE Spoken Language Technology Workshop (SLT)}, pages
  324--329. IEEE, 2014.

\bibitem{williams2014dialog}
Jason~D Williams, Matthew Henderson, Antoine Raux, Blaise Thomson, Alan Black,
  and Deepak Ramachandran.
\newblock The dialog state tracking challenge series.
\newblock {\em AI Magazine}, 35(4):121--124, 2014.

\bibitem{williams2016dialog}
Jason Williams, Antoine Raux, and Matthew Henderson.
\newblock The dialog state tracking challenge series: A review.
\newblock {\em Dialogue \& Discourse}, 7(3):4--33, 2016.

\bibitem{el-asri-etal-2017-frames}
Layla El~Asri, Hannes Schulz, Shikhar Sharma, Jeremie Zumer, Justin Harris,
  Emery Fine, Rahul Mehrotra, and Kaheer Suleman.
\newblock {F}rames: a corpus for adding memory to goal-oriented dialogue
  systems.
\newblock In {\em Proceedings of the 18th Annual {SIG}dial Meeting on Discourse
  and Dialogue}, pages 207--219. Association for Computational Linguistics,
  August 2017.

\bibitem{eric-etal-2017-key}
Mihail Eric, Lakshmi Krishnan, Francois Charette, and Christopher~D. Manning.
\newblock Key-value retrieval networks for task-oriented dialogue.
\newblock In {\em Proceedings of the 18th Annual {SIG}dial Meeting on Discourse
  and Dialogue}, pages 37--49. Association for Computational Linguistics,
  August 2017.

\bibitem{Shah2018BuildingAC}
Pararth Shah, Dilek~Z. Hakkani-T{\"u}r, G{\"o}khan T{\"u}r, Abhinav Rastogi,
  Ankur Bapna, Neha~Nayak Kennard, and Larry Heck.
\newblock Building a conversational agent overnight with dialogue self-play.
\newblock {\em ArXiv}, abs/1801.04871, 2018.

\bibitem{zang2020multiwoz}
Xiaoxue Zang, Abhinav Rastogi, Srinivas Sunkara, Raghav Gupta, Jianguo Zhang,
  and Jindong Chen.
\newblock {M}ulti{WOZ} 2.2 : A dialogue dataset with additional annotation
  corrections and state tracking baselines.
\newblock In {\em Proceedings of the 2nd Workshop on Natural Language
  Processing for Conversational AI}, pages 109--117. Association for
  Computational Linguistics, July 2020.

\bibitem{li-etal-2017-dailydialog}
Yanran Li, Hui Su, Xiaoyu Shen, Wenjie Li, Ziqiang Cao, and Shuzi Niu.
\newblock {D}aily{D}ialog: A manually labelled multi-turn dialogue dataset.
\newblock In {\em Proceedings of the Eighth International Joint Conference on
  Natural Language Processing (Volume 1: Long Papers)}, pages 986--995. Asian
  Federation of Natural Language Processing, November 2017.

\bibitem{ritter-etal-2010-unsupervised}
Alan Ritter, Colin Cherry, and Bill Dolan.
\newblock Unsupervised modeling of {T}witter conversations.
\newblock In {\em Human Language Technologies: The 2010 Annual Conference of
  the North {A}merican Chapter of the Association for Computational
  Linguistics}, pages 172--180. Association for Computational Linguistics, June
  2010.

\bibitem{lison2016opensubtitles2016}
Pierre Lison and J{\"o}rg Tiedemann.
\newblock {O}pen{S}ubtitles2016: Extracting large parallel corpora from movie
  and {TV} subtitles.
\newblock In {\em Proceedings of the Tenth International Conference on Language
  Resources and Evaluation ({LREC}'16)}, pages 923--929. European Language
  Resources Association (ELRA), May 2016.

\bibitem{Danescu-Niculescu-Mizil+Lee:11a}
Cristian Danescu-Niculescu-Mizil and Lillian Lee.
\newblock Chameleons in imagined conversations: A new approach to understanding
  coordination of linguistic style in dialogs.
\newblock In {\em Proceedings of the Workshop on Cognitive Modeling and
  Computational Linguistics, ACL 2011}, 2011.

\bibitem{yang2019friendsqa}
Zhengzhe Yang and Jinho~D Choi.
\newblock Friendsqa: Open-domain question answering on tv show transcripts.
\newblock In {\em Proceedings of the 20th Annual SIGdial Meeting on Discourse
  and Dialogue}, pages 188--197, 2019.

\bibitem{AMIcorpus}
Iain Mccowan, J~Carletta, Wessel Kraaij, Simone Ashby, S~Bourban, M~Flynn,
  M~Guillemot, Thomas Hain, J~Kadlec, V~Karaiskos, M~Kronenthal, Guillaume
  Lathoud, Mike Lincoln, Agnes Lisowska~Masson, Wilfried Post, Dennis Reidsma,
  and P~Wellner.
\newblock The ami meeting corpus.
\newblock {\em Int'l. Conf. on Methods and Techniques in Behavioral Research},
  01 2005.

\bibitem{kim-etal-2023-soda}
Hyunwoo Kim, Jack Hessel, Liwei Jiang, Peter West, Ximing Lu, Youngjae Yu, Pei
  Zhou, Ronan Bras, Malihe Alikhani, Gunhee Kim, Maarten Sap, and Yejin Choi.
\newblock {SODA}: Million-scale dialogue distillation with social commonsense
  contextualization.
\newblock In {\em Proceedings of the 2023 Conference on Empirical Methods in
  Natural Language Processing}, pages 12930--12949. Association for
  Computational Linguistics, December 2023.

\bibitem{chen-etal-2023-places}
Maximillian Chen, Alexandros Papangelis, Chenyang Tao, Seokhwan Kim, Andy
  Rosenbaum, Yang Liu, Zhou Yu, and Dilek Hakkani-Tur.
\newblock {PLACES}: Prompting language models for social conversation
  synthesis.
\newblock In {\em Findings of the Association for Computational Linguistics:
  EACL 2023}, pages 844--868. Association for Computational Linguistics, May
  2023.

\bibitem{myat2024framework}
Kaung Myat~Kyaw and Jonathan Hoyin~Chan.
\newblock A framework for synthetic audio conversations generation using large
  language models.
\newblock {\em arXiv e-prints}, pages arXiv--2409, 2024.
\newblock in press.

\bibitem{ding2023gpt3gooddataannotator}
Bosheng Ding, Chengwei Qin, Linlin Liu, Yew~Ken Chia, Boyang Li, Shafiq Joty,
  and Lidong Bing.
\newblock Is {GPT}-3 a good data annotator?
\newblock In {\em Proceedings of the 61st Annual Meeting of the Association for
  Computational Linguistics (Volume 1: Long Papers)}, pages 11173--11195.
  Association for Computational Linguistics, July 2023.

\bibitem{pangakis2023automatedannotationgenerativeai}
Nicholas Pangakis, Samuel Wolken, and Neil Fasching.
\newblock Automated annotation with generative ai requires validation, 2023.

\bibitem{he2024annollmmakinglargelanguage}
Xingwei He, Zhenghao Lin, Yeyun Gong, A-Long Jin, Hang Zhang, Chen Lin, Jian
  Jiao, Siu~Ming Yiu, Nan Duan, and Weizhu Chen.
\newblock {A}nno{LLM}: Making large language models to be better crowdsourced
  annotators.
\newblock In {\em Proceedings of the 2024 Conference of the North American
  Chapter of the Association for Computational Linguistics: Human Language
  Technologies (Volume 6: Industry Track)}, pages 165--190. Association for
  Computational Linguistics, June 2024.

\bibitem{pangakis2024knowledgedistillationautomatedannotation}
Nicholas Pangakis and Sam Wolken.
\newblock Knowledge distillation in automated annotation: Supervised text
  classification with {LLM}-generated training labels.
\newblock In {\em Proceedings of the Sixth Workshop on Natural Language
  Processing and Computational Social Science (NLP+CSS 2024)}, pages 113--131.
  Association for Computational Linguistics, June 2024.

\bibitem{Devlin2019BERTPO}
Jacob Devlin, Ming-Wei Chang, Kenton Lee, and Kristina Toutanova.
\newblock Bert: Pre-training of deep bidirectional transformers for language
  understanding.
\newblock In {\em North American Chapter of the Association for Computational
  Linguistics}, 2019.

\bibitem{choi-etal-2018-quac}
Eunsol Choi, He~He, Mohit Iyyer, Mark Yatskar, Wen-tau Yih, Yejin Choi, Percy
  Liang, and Luke Zettlemoyer.
\newblock {Q}u{AC}: Question answering in context.
\newblock In {\em Proceedings of the 2018 Conference on Empirical Methods in
  Natural Language Processing}, pages 2174--2184. Association for Computational
  Linguistics, October-November 2018.

\bibitem{rodriguez2020curiosity}
Pedro Rodriguez, Paul Crook, Seungwhan Moon, and Zhiguang Wang.
\newblock Information seeking in the spirit of learning: a dataset for
  conversational curiosity.
\newblock In {\em Empirical Methods in Natural Language Processing}, 2020.

\bibitem{kwiatkowski2019natural}
Tom Kwiatkowski, Jennimaria Palomaki, Olivia Redfield, Michael Collins, Ankur
  Parikh, Chris Alberti, Danielle Epstein, Illia Polosukhin, Jacob Devlin,
  Kenton Lee, et~al.
\newblock Natural questions: a benchmark for question answering research.
\newblock {\em Transactions of the Association for Computational Linguistics},
  7:453--466, 2019.

\bibitem{Li2018MicrosoftDC}
Xiujun Li, Sarah Panda, Jingjing Liu, and Jianfeng Gao.
\newblock Microsoft dialogue challenge: Building end-to-end task-completion
  dialogue systems.
\newblock {\em ArXiv}, abs/1807.11125, 2018.

\bibitem{fu2016tie}
Liye Fu, Cristian Danescu-Niculescu-Mizil, and Lillian Lee.
\newblock Tie-breaker: Using language models to quantify gender bias in sports
  journalism.
\newblock In {\em Proceedings of the IJCAI workshop on NLP meets Journalism},
  2016.

\bibitem{chawla2021casino}
Kushal Chawla, Jaysa Ramirez, Rene Clever, Gale Lucas, Jonathan May, and
  Jonathan Gratch.
\newblock Casino: A corpus of campsite negotiation dialogues for automatic
  negotiation systems.
\newblock In {\em Proceedings of the 2021 Conference of the North American
  Chapter of the Association for Computational Linguistics: Human Language
  Technologies}, page 3167–3185, 2021.

\bibitem{zhang2016conversational}
Justine Zhang, Ravi Kumar, Sujith Ravi, and Cristian Danescu-Niculescu-Mizil.
\newblock Conversational flow in {O}xford-style debates.
\newblock In {\em Proceedings of the 2016 Conference of the North {A}merican
  Chapter of the Association for Computational Linguistics: Human Language
  Technologies}, pages 136--141. Association for Computational Linguistics,
  June 2016.

\bibitem{jandaghi-etal-2024-faithful}
Pegah Jandaghi, Xianghai Sheng, Xinyi Bai, Jay Pujara, and Hakim Sidahmed.
\newblock Faithful persona-based conversational dataset generation with large
  language models.
\newblock In {\em Proceedings of the 6th Workshop on NLP for Conversational AI
  (NLP4ConvAI 2024)}, pages 114--139. Association for Computational
  Linguistics, August 2024.

\bibitem{Wu_AutoGen_Enabling_Next-Gen_2024}
Qingyun Wu, Gagan Bansal, Jieyu Zhang, Yiran Wu, Beibin Li, Eric Zhu, Li~Jiang,
  Shaokun Zhang, Xiaoyun Zhang, Jiale Liu, Ahmed~Hassan Awadallah, Ryen~W
  White, Doug Burger, and Chi Wang.
\newblock {AutoGen: Enabling Next-Gen LLM Applications via Multi-Agent
  Conversation Framework}, 2024.

\end{thebibliography}

\vspace{12pt}

\newpage
\onecolumn
\appendix

\begin{appendices}
\setcounter{table}{0}
\section{Data Collection}\label{appendix:DataCollection}
The following table shows a break down of the training data that we utilized for fine-tuning the intent-based filter for multi-label classification.
\begin{table}[h!]
\centering
\caption{\centering Train Dataset Summary}
\begin{tabular}{|l|r|r|r|r|r|}
\hline
\textbf{Dataset} & \textbf{Total} & \textbf{Requires Action} & \textbf{Not Requiring Action} & \textbf{Requires Information} & \textbf{Not Requiring Information}  \\
\hline
AMI – 1 min chunks & 4044 & 1860 & 2184 & 1608 & 2436  \\
FriendsQA & 2996 & 1749 & 1247 & 290 & 2706  \\
IQ2 Corpus & 2978 & 901 & 2077 & 1417 & 1561  \\
Movie Corpus & 57080 & 10693 & 46387 & 5513 & 51567 \\
CaSiNo & 980 & 972 & 8 & 11 & 969  \\
Tennis Interviews & 1683 & 1478 & 205 & 576 & 1107 \\
Taxi Conversations & 2890 & 2552 & 338 & 177 & 2713  \\
QuAC & 11567 & 0 & 11567 & 11567 & 0  \\
MultiWOZ & 7933 & 6657 & 1276 & 7159 & 774  \\
Curiosity Dialogues & 10283 & 860 & 9423 & 10256 & 27  \\
Synthetic Situations & 2827 & 2540 & 287 & 287 & 2540  \\
Synthetic Information Seeking & 19938 & 17842 & 2096 & 19935 & 3  \\
Synthetic Multi-turn Assistant & 4935 & 4935 & 0 & 144 & 4791  \\
In-house Assistant Commands & 147471 & 147471 & 0 & 54 & 147417 \\
Natural Questions & 61503 & 0 & 61503 & 61503 & 0  \\
\hline
\textbf{Total} & 339108 & 200510 & 138598 & 120497 & 218611 \\
\textbf{\%Synthetic} & 8.17\% & 12.63\% & 1.72\% & 16.9\% & 3.35\% \\
\hline
\end{tabular}

\label{tab:train_dataset_summary}
\end{table}

\begin{table}[h!]
\centering
\caption{\centering Validation Dataset Summary}
\begin{tabular}{|l|r|r|r|r|r|}
\hline
\textbf{Dataset} & \textbf{Total} & \textbf{Requires Action} & \textbf{Not Requiring Action} & \textbf{Requires Information} & \textbf{Not Requiring Information} \\
\hline
AMI - 1 min chunks & 500 & 253 & 247 & 187 & 313 \\
Movie Corpus & 197 & 98 & 99 & 16 & 181 \\
QuAC & 1000 & 0 & 1000 & 1000 & 0 \\
MultiWoZ & 973 & 906 & 67 & 886 & 87 \\
Curiosity Dialogues & 1287 & 105 & 1182 & 1282 & 5 \\
Synthetic Situations & 100 & 50 & 50 & 12 & 88 \\
\hline
\end{tabular}

\label{tab:val_dataset_summary}
\end{table}

\begin{table}[h!]
\centering
\caption{\centering Test Dataset Summary}
\begin{tabular}{|l|r|r|r|r|r|}
\hline
\textbf{Dataset} & \textbf{Total} & \textbf{Requires Action} & \textbf{Not Requiring Action} & \textbf{Requires Information} & \textbf{Not Requiring Information} \\
\hline
AMI - 1 min chunks & 502 & 243 & 259 & 194 & 308 \\
Movie Corpus & 193 & 96 & 97 & 17 & 176 \\
Curiosity Dialogues - 0 & 1187 & 86 & 1101 & 1185 & 2 \\
Curiosity Dialogues - 1 & 1287 & 115 & 1172 & 1286 & 1 \\
MultiWOZ & 984 & 924 & 60 & 891 & 93 \\
Synthetic Situations & 80 & 50 & 30 & 8 & 72 \\
Synthetic Multi-turn Assistant & 1113 & 1113 & 0 & 23 & 1090 \\
In-house Assistant Commands & 17258 & 17258 & 0 & 417 & 16841 \\
Balanced Test & 1866 & 999 & 867 & 1138 & 728 \\
\hline
\end{tabular}

\label{tab:test_dataset_summary}
\end{table}

\end{appendices}
\end{document}